# On the influence of intelligence in (social) intelligence testing environments

Javier Insa-Cabrera
jinsa@dsic.upv.es
DSIC, UPV

José-Luis Benacloch-Ayuso
jobeay@fiv.upv.es
ETSINF, UPV

José Hernández-Orallo
jorallo@dsic.upv.es
DSIC, UPV

November 12, 2018

**Abstract**

This paper analyses the influence of including agents of different degrees of intelligence in a multiagent system. The goal is to better understand how we can develop intelligence tests that can evaluate social intelligence. We analyse several reinforcement algorithms in several contexts of cooperation and competition. Our experimental setting is inspired by the recently developed Darwin-Wallace distribution.

## 1 Introduction

Dating back from the late nineties, but with a stronger momentum recently, we can find several works [1, 10, 3, 14, 8, 2] addressing the problem of measuring agent intelligence in a principled and general way. With the common thing of using notions taken from (algorithmic) information theory, MML and two-part compression, Kolmogorov complexity and Solomonoff priors (see [15] for proper definitions of all these notions), some of these works present definitions and tests to evaluate agent intelligence. One important feature in some of these tests is that the complexity of a problem, task or environment can be derived from its Kolmogorov complexity. This allows for the application of the setting to many different fields in artificial intelligence, including inductive or deductive tasks [4, 6]: given any task or problem, we can derive its intrinsic complexity and use it as a measure of difficulty.

A prototype of the *universal* test introduced in [8] has been used to evaluate several agents, including humans and reinforcement learning agents. Some preliminary results of this evaluation [11, 12] show that the setting is able to compare and evaluate different kinds of agents, but it fails at placing them on the same scale, since humans usually get similar scores to other relatively simple agents.

One possible explanation for these results is that it is not usual to find other agents in the test, so social intelligence is rarely measured. The question, however, is what agents should be introduced in the test. This is related to the question of evaluating intelligence with games (also suggested in [8]), where the difficulty is not only given by the complexity of the game, but from the opponent's intelligence. Fig. 1 shows this situation. This leads to a circular problem: we need to know the opponent's intelligence first in order to know the complexity of the problem.

One recent proposal to overcome this problem is turning this circularity into a recursion. The Darwin-Wallace distribution [9] establishes a distribution of agents based on an evolutionary process. The first 'generation' just uses a Solomonoff prior over agents, with very simple agents predominating. These agents are set to interact in a random environment. The second generation is constructed by selecting the agents according to their performance. The result of this (apparently intractable) evolutionary process is a distribution of mind forms, i.e., a distribution of agents. The higher the generation $i$ is, the more socially intelligent their agents should be or, in other words, the more demanding the 'society' will be, in the sense that competing and collaborating with other socially intelligent agents requires social intelligence. Note that this does not mean that minds have to evolve as in a true evolutionary process. This is, of course, related to evolutionary game theory [16, 17, 22], where the 'game' is now an intelligence test. In addition, the 'game rules' can change at any time by switching to virtually any environment and the replicator rules are the result of the intelligence test with some probability of agent reintroduction based on universal distributions.

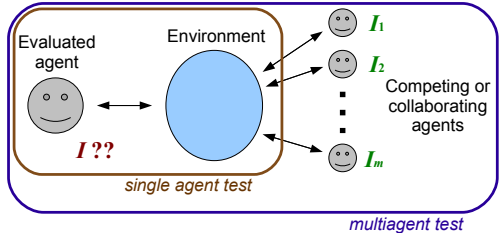

Figure 1: A multiagent intelligence test compared to a single agent intelligence test. In a multiagent (social) intelligence test, other agents also interact (and become integral part) of the environment. In order to assess the intelligence of the evaluated agents, we need to know the intelligence of the other agents.

The previous proposal can be revealing about the kind of environments where (social) intelligence excels, but it has many implementation issues. Nonetheless, it gives some clues about how social intelligence can be measured and how the other agents can be chosen. In fact, it is suggested in [9] that intelligence tests could be used to make this choice of agents, using off-the-shelf algorithms in AI.

In this paper, we perform some experiments on a general intelligence test setting in order to examine the way in which simple competitive and cooperative scenarios may have a big impact on the performance of some simple agents. This is a necessary first step to analysing whether an intelligence test can be used effectively as a 'game'. This is also crucial to determine how social environments affect the results obtained by different agents in order to get more information about how to approximate the Darwin-Wallace distribution. We will use very simple reinforcement learning (RL) algorithms: SARSA [18], Q-learning [21] and QV-learning [23]. The goal of the paper is not showing how these three algorithms behave nor comparing them. We just use them as off-the-shelf agents which can learn from an environment to see how performance is affected by the introduction of more agents in an environment. Rather, the true goal of the paper is analysing the behaviour of intelligence tests when environments are populated with agents, and how this affects the results of the evaluated agent. We will examine several scenarios, some with competition and some with cooperation. We focus on how rewards must be changed from the scenario where only one agent is evaluated and no other agents compete, to the scenario where several agents compete for the rewards and the scenario where the agents accumulate (or average) all the rewards of their team.

The paper is organised as follows. Section 2 reviews the notion of universal intelligence test and how the Darwin-Wallace distribution can be useful to turn them into social intelligence tests. Section 3 makes the extension by modifying the environments and the reward system for competing and cooperating scenarios. The following sections 4, 5, 6 and 7 perform and discuss the experiments of the different scenarios. A more comprehensive discussion of results and implications is found in section 8. Finally, section 9 closes the paper with the conclusions and future work.

## 2 Universal intelligence tests and social intelligence

Social intelligence has been defined in many ways in psychology and cognition, but it can be just worded, in the terminology of agents, as the ability to perform well in the context of other agents. One problem of this definition is that we have to be more precise about what the 'other agents' are. If these other agents are bacteria or sea sponges, then our intuitive notion of social intelligence does not work well, because working well in the context of other agents with low intelligence is not necessarily related to social intelligence, as we know it. In psychometrics and human cognition, social intelligence clearly sets these other agents as other humans. But what about artificial agents? If we use a society of dull agents, the abilities which may be required to perform well might be very different to those which are required if we introduce an agent into, e.g., a society of humans.

The difference between social intelligence and general intelligence is that in the latter an agent could perform well if it were able to solve non-social tasks, such as escaping from a maze, solving a puzzle or predicting the next number in a series. On the contrary, social intelligence implies that tasks involve competing and collaborating with other agents.

One approach for measuring intelligence is to take a diverse selection of tasks of different complexity and to measure agent performance over this selection. However, several issues arise here. The selection of tasks must be unbiased, i.e., we cannot measure intelligence by solely measuring addition ability, for instance. One approach is to consider all possible (computable) tasks, as defined by a universal Turing machine. In order to link performance to any possible task we can use the notion of rewards. This leads to interactive scenarios, which can be well represented by (discrete) environments, very much like the observaton-action-reward environments which are typical in reinforcement learning [20, 5] and other areas in agent theory. Finally, we need to assess the complexity of each task in order to make a proper choice of tasks which capture a wide range of difficulty and, therefore, can suit the agent's level of intelligence. These issues have been addressed in [1, 10, 3, 14, 8].

An evaluation session or episode is very similar to an experiment in reinforcement learning. An agent can interact with an environment by performing actions and receives observations and rewards. The performance of the agent is evaluated as the average reward throughout the whole session (the average reward of all the interactions). So, a test is a set of evaluation sessions over different environments.

In this context, [8] introduces the idea of universal test, a test which is conceived to be feasibly applicable to any kind of agent: humans, non-human animals, artificial agents, including hybrids and communities, of any degree of intelligence and speed. The test is based on a set of environments as in [14]. The complexity of tasks is derived in a formal way, using notions from Kolmogorov complexity, and then used to define an adaptive procedure. The focus of [8] is set on feasibility, highlighting that a sample of environments must be chosen carefully. This choice can be done as a postprocess (sieving the environments which, empirically, follow some properties) or it can be determined a priori, by defining a proper environment class following some appropriate properties. In this sense, choosing an environment class which considers any computable environment might seem very unbiased (although it depends on the reference machine chosen) but it is impractical because of several reasons. First, according to [8] we need environments to be discriminative, so the behaviour of the agent has impact on its rewards, i.e., the environments should not lead to deadends, because they lead to states where anything the agent does is useless to change its rewards. Additionally, we want environments to be balanced, i.e., that a random agent scores 0 on average, where rewards go from $-1$ to $1$.

Keeping these conditions in mind while still trying to be general enough to consider any possible behaviour requires a trade-off. In [7], a hopefully unbiased environment class (called $\Lambda$) is introduced, as a class of environments composed of spaces and agents with universal descriptive (Turing-complete) power. Originally, only two agents (apart from the evaluated agent) were used in the environments, but since their behaviour is generated by a universal distribution using Markov algorithms (a Turing-complete rewriting language), their sophistication is really low and it was very difficult to find any social behaviour originating from them, and therefore, any social behaviour in the environments. The first evaluations using these tests [11] show that they work well at evaluating very different agents (humans, and RL algorithms), but they do not properly reflect their supposed difference in intelligence. Many possible explanations are suggested in [11] for this phenomenon, with incremental knowledge acquisition and social intelligence being two of the abilities which this test is not giving enough importance.

In order to address the second issue, one direct option is to define more social environments. But if we just define a social environment as any environment where we include other agents, we may have counter-intuitive situations. For instance, if the reactivity and intelligence of the other agents are those of a stone, we cannot properly say that this is a social environment. The question of which agents are introduced becomes more important if we want to use the environment as a testbed, since the results of the evaluated agent will depend on the abilities of the other agents. This is illustrated in Fig. 1. This is well-known in the area of games and multiagent systems, but it is not so common in the area of (social) intelligence tests. The question is what criteria we can use to introduce the other agents and how can we measure their (social) intelligence in advance. Just selecting opponents at random would lead to a relativist view of intelligence, where we could compare two agents (on one or more classes of tasks), but we would not be able to construct a comprehensive and universal scale. In [9], instead of incorporating other agents in an ad-hoc way, the authors stick to the fundamental notions of the previous test proposals, i.e. there must be a formal way to determine which agents are introduced in a social environment. As [9] states: "The basic idea is straightforward: intelligence is the result of evolution through millions of generations interacting with other live beings. Thus we define intelligence in this context, interacting with other agents of similar intelligence". From this idea they formalise the so-called Darwin-Wallace distribution for agents and environments.

Informally, the Darwin-Wallace distribution requires the notion of multiagent environment, an environment $\mu$ which has its rewards, actions and observations as usual, but allows the 'introduction' of any number $m$ of agents $\pi_1 \ldots \pi_m$. From here, the Darwin-Wallace distribution is defined recursively according to a level or generation $i$:

1. The first generation is denoted by $i = 0$.

2. The distribution of environments for $i = 0$ is chosen as a universal distribution of multiagent environments over a universal Turing machine $U_e$. Formally, $p_E(\mu) := 2^{-K_{U_e}(\mu)}$, where $K()$ refers to Kolmogorov complexity. This is the probability of the base multi-agent environment not considering the agents.

3. The distribution of agents for $i = 0$ is also chosen as a universal distribution of agents over a UTM $U_a$. Formally, $p_A^0(\pi) := 2^{-K_{U_a}(\pi)}$.

4. Everything is put together in a distribution for multiagent environments $\sigma$ with $m$ agents: $p_S^i(\sigma) = p_S^i(\langle \mu, \pi_1, \pi_2, ..., \pi_m \rangle) := p_E(\mu) \times \prod_{j=1}^{m} p_A^i(\pi^j)$.

5. Agents interact with the environment for several steps for a generation $i$. Then $i$ is incremented.

6. Two mechanisms are included to update environments and agents. For environments, a probability of replacement $c$ is defined, which, in case, replaces the environment by another one from the same $p_E(\mu)$. The probability of dying $d$ for an agent is defined as a function of the past rewards. The agent is replaced with another agent using $p_A^{i-1}(\pi)$. This makes up a new agent distribution $p_A^i(\pi)$.

7. This resulting agent probability distribution $p_A^i$ is used to define $p_S^i$, using the formula in step 4.

The use of this distribution has many issues. First, it is a theoretical construct which might be useful to understand the kind of environments where intelligence is appropriate, since it implicitly gives a definition of (social) intelligence as the ability of performing well in a variety of multiagent environments with a variety of agents. Second, this distribution could be used for the construction of social intelligence tests, just sampling from the distribution. Third, and recursively, the way in which this distribution can be approximated is precisely by the use of intelligence tests, where human-made agents can be incorporated into the environments, provided we have been able to assess their intelligence first.

Following this last issue, we need to develop intelligence tests in multiagent scenarios. In particular, we need to adapt the existing intelligence test proposals to a multiagent setting. This is what we do below.

## 3 Extending an intelligence test to consider several agents

The first intelligence tests based on the theory developed in [8] have been based on the environment class $\Lambda$, introduced in [7]. We summarise this class first and then we see the way in which tests can be constructed using this class, as done in [11].

This environment class considers a space which is composed of a directed labelled graph, where vertices are cells and arrows are actions. The graph is selected to be strongly connected (all cells are reachable from any other cell). Cells can contain agents. Every environment must include at least three agents: the evaluated agent, and two special agents *Good* and *Evil*. *Good* and *Evil* are not generally reactive, so, if no further agent is included, the environment cannot be considered a proper multiagent system. Actions allow the evaluated agent (and other agents) to move in the space. Observations show the cell contents. Rewards are rational numbers in the interval $[-1, 1]$ and are generated by the agents *Good* and *Evil*, which leave rewards in the cells they visit. Before dropping the new reward, any old reward value is previously erased from the cell. Rewards do not stay unaltered in the cell forever. If a reward in a cell is eaten by any agent (including *Good* and *Evil*) because the agent steps into or stays in the cell, the reward disappears. While rewards are not eaten, their value is divided by 2 for each iteration. This has the effect of seeing *Good* and *Evil* as agents which leave a reward wake as they move. *Good* and *Evil* have the same behaviour (they follow the same pattern) except for the sign of the reward ($+$ for *Good*, $-$ for *Evil*). This makes *Good* and *Evil* symmetric, which ensures that the environment is balanced (random agents score 0 on average) [8]. *Good* and *Evil* are initially placed randomly in different cells. *Good* and *Evil* cannot share a cell, and when both move to the same cell at the same time, the conflict is resolved randomly. For more details of the environment class $\Lambda$, see [7]. Below, we follow with some simplifications taken in order to construct test prototypes based on the environment class.

A session or episode is the evaluation of an agent against an environment for a number of iterations. A test is just a set of episodic evaluations. Consequently, in order to generate a test we need a sample of environments. Each environment is then randomly generated as follows. Spaces are generated by first determining the number of cells $n_c$ (in this paper we set $n_c = 9$ but other configurations have been explored in [11]). The number of possible actions $n_a$ is defined with a geometric distribution between 2 and $n_c$. The connections between cells are determined using a uniform distribution for each cell, among the possible actions and cells. We consider the possibility that some actions do not lead to any cell, so these actions have no effect. The sequence of actions for *Good* and *Evil* is defined by using a uniform distribution for each element in the sequence, and a geometric distribution to determine whether to stop the sequence, by using a probability of stopping ($p_{stop} = 1/100$). The agents *Good* and *Evil* take one action from the sequence and execute it. Then they take the second action, and so on, for each step in the system. When the actions are exhausted, the sequence is started all over again. If an action is not allowed at a particular cell, the agent does not move. This is a simplification by the prototype implemented by [11] of the originally Turing-complete behaviour of *Good* and *Evil*, and it also makes them non-reactive.

Once an environment has been constructed, evaluation is performed in the following way. Initially, each agent is randomly (using a uniform distribution) placed in a cell. Then, we let *Good*, *Evil*, the evaluated agent and any other agents in the space interact for a certain number of steps, i.e., a session. For a session we average the rewards, thus giving a score of the agent in the environment.

This configuration has been used to evaluate agents separately in [11]. Fortunately, this configuration makes it relatively easy to extend this single agent evaluation setting into a multiagent setting. It is just direct to include more agents in the environment. We do not need any further constraint. Consequently, agents can move freely to other cells independently of whether they are occupied or not by other agents. In other words, common agents can share a cell. The behaviour of *Good* and *Evil* is modified slightly in such a way that they avoid stepping into a cell where any other agent is located. This re-introduces some degree of reactivity (with respect to the prototype in [11]), even in the single agent case (remember that we do not count *Good* and *Evil* as proper agents).

There are many possible ways of introducing cooperation and competition, which may lead to different experimental results, some of them similar to what has been previously studied in the literature. In this paper we do not want to evaluate these choices, but to analyse *how the degree of intelligence of the agents in a social environment affects the role of cooperation and competition*. The ultimate goal is to shed some light on whether environments become difficult when many agents are introduced (independently of their intelligence) or become difficult (and socially challenging) when other intelligent agents are introduce. These findings are necessary if we aim at measuring social intelligence.

The first question when several agents are introduced in the space is how the rewards are shared among them. This is relative easy to solve, as well. If two or more agents share a cell, each one receives the reward in the cell, divided by the number of agents in the cell. This clearly sets a purely competitive scenario.

A second, relatively more difficult, question is how we can deal with cooperation. It can be argued that alliances can be established in a purely competitive scenario in such a way that cooperation might be valuable. However, this would be a rare situation. Consequently, it is better to figure out a modification of the reward system such that cooperation is valuable per se. The easiest way of making this setting purely cooperative is by just putting all the rewards in the same bag. With this, one should not be concerned about getting some reward if some other agent is able to get it instead. What matters is the overall result.

We can of course move between these two points by using the notion of team. All the members of a team put their rewards in the same bag. If we have two teams, both teams should compete against each other. This arrangement is usual in games and economics.

Now we are ready to see what happens in a single agent intelligence test when we turn it into a multiagent test. But, before that, we need to determine the agents that we will use for the experiment. The agents are:

- Oracle: this agent 'foresees' the cell where *Good* will be at the next step. If this cell is one of the neighbouring cells then it moves to that cell. Otherwise, it goes to the adjacent cell that, in the next iteration, will have the highest reward.

- Trivial Follower: this is an agent which looks at the neighbouring cells to see whether *Good* is in one of them. If it finds it, then it moves to that cell. Otherwise, it makes a random move trying to avoid *Evil*.

- Random: a random agent is just an agent which chooses randomly among the available actions using a uniform distribution.

- Q-learning: the most common algorithm in reinforcement learning, as explained in [21] and [20]. We use the description of cell contents as a state.

- SARSA [18]. This is a well-known variant of Q-learning which also takes the future state and action into account when updating the Q-value.

- QV-learning [23]. This is also a variant of Q-learning which partially resembles ActorCritic methods. In contrast to Q-learning and SARSA, it keeps track of two functions, the Q-function and the value function $V$. The eligibility trace is not implemented.

The three latter algorithms will be referred to as RL agents. These three algorithms have many parameters. In order to have a consistent view of the experiments, the parameters for all the algorithms (*learning rate* $\alpha$, *discount factor* $\gamma$ and others) were fine-tuned on the single agent scenarios, by using 1,000 sessions for each parameter variation, totalling a huge number of experiments to set the optimal parameters. These parameter-setting experiments were performed before starting with the experiments which follow below.

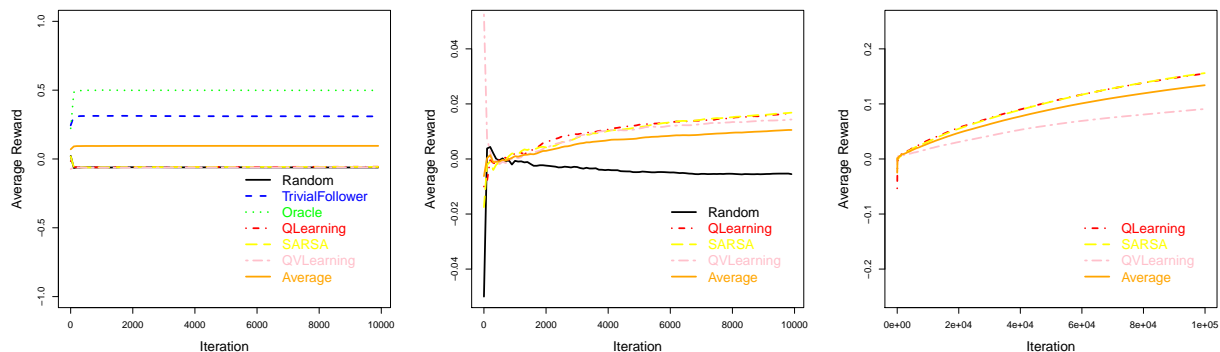

Figure 2: Competition scenarios. Left: The 6 evaluated agents are evaluated. Middle: The 3 RL agents along with the random agent. Right: The 3 RL agents without the random agent. 100 environments each.

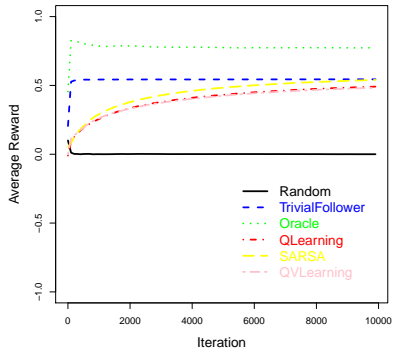

Figure 3: Isolated scenario. The 6 evaluated agents are evaluated independently over the same test. Results after 10,000 iterations with 100 environments.

## 4 Evaluating agents isolatedly

We start our experiments with the scenario where agents are just taken and evaluated isolatedly. This is the same setting as in [11], with the only (minor) difference that *Good* and *Evil* are slightly reactive because they try to avoid sharing a cell with other agents. In addition, we will just restrict the evaluation to environments with nine cells.

The result of Fig. 3 is clear (and consistent with the results in [11]). The oracle gets very good results (but not optimal since there are some stochastic behaviour it cannot anticipate, and also because the oracle does not plan further than the immediately best action) and the trivial follower stabilises slightly above 0.5. The random agent has an average reward of 0, as predicted by the theory. The three RL agents are very slow learners and only get closer to the trivial follower after 10,000 iterations. Their behaviour is similar and the differences are small.

## 5 Evaluating agents in a competitive scenario

More interesting things can be observed when we switch to the competitive scenario. Remember that here, all the agents are located in the environment at the same time, and they compete for rewards. Note that the sum of rewards is bounded, so if one agent performs well (e.g. better than $1/6 = 0.167$), then the others will get necessary worse, as there will remain less positive rewards and they can still get the negative ones.

This is precisely what Fig. 2 (left) shows. If we look at the random agent, we see that it gets a value which is even lower than 0, since most of the positive rewards are eaten by the other agents, and it may only collect negative rewards. It is not much lower than 0, but this can be explained because there are nine cells and rewards are divided by two for each iteration, so a random walk will frequently find cells with almost no (negative) reward. We also see that the oracle gets good results, but since it has to compete with other agents, its results are worse than in the single agent scenario. A similar thing happens to the trivial follower. However, it is relatively surprising to see how the RL algorithms collapse, behaving equally bad as a random agent. One alleged reason might be that it is futile to compete with the oracle. However, this is not true, since the trivial follower is able to share some rewards with the oracle and to get almost all the reward that the oracle loses. The real reason is that having so many agents dramatically increases the perception combinations and, consequently, the state tables these RL algorithms use.

In order to confirm this, we have repeated the experiment without the oracle and the trivial follower. Now, in Fig. 2 (middle) we see that even without good competitors, RL algorithms have a very poor result (not reaching 0.02 in 10,000 iterations). This is consistent with the previous explanation.

Finally, in order to further confirm that the problem is the state space, we remove the random agent (which, given its random behaviour, can be considered noise which contributes to this huge state space), and we only leave the three RL agents. We also increase the number of iterations to 100,000. This is shown in Fig. 2 (right). Things improve slightly and, in the very long term, Q-learning and SARSA get close to 0.2, while QV-learning lagging behind around 0.1. Even the presence of only two other agents makes their matrices so big that they require more than 100,000 iterations to derive their $Q$-values accurately.

Apart from the comparative results, we clearly see that performance depends, as expected, on the other agents' policies but more especially on the ability of digesting the state space, and how much noise (from the random agent) can be handled. Finally, it is interesting to mention that one of the things that have been lost when other agents are introduced is the notion of balancedness, since a random agent will typically score worse than 0.

## 6 Evaluating agents in a cooperative scenario

The next scenario we want to explore is when the 6 agents are prompted to cooperate. This is done by putting all the rewards in the same bag, so the agents just see the reward as the average reward of all the agents. The results for all the agents are shown in Fig. 4 (left). The oracle, the trivial follower and the random agent cannot change their behaviour, so it is clear that their results should be similar to those in Fig. 2 (left) if the other agents do not make a dramatic improvement. But this is not the case. RL agents were already lost by the state space in the competitive case, and they are equally lost in the cooperative case, since the length of the observations is the same.

Fig. 4 (middle) changes from Fig. 2 (middle). How can it be that moving form a competitive to a cooperative case, we get worse average results? Should not it be the other way round? Here the explanation is a little bit more convoluted. The problem of cooperation is the way we assign rewards. Since the reward they receive is the average of the rewards of all the agents, it is much more difficult for them to determine the goodness of the actions, since rewards are affected by other agents' movements.

This explanation is only part of the story if we take a look at Fig. 4 (right), and we compare it to Fig. 4 (middle) and to Fig. 2 (middle). In this case, where the random agent has been removed, the agents require a large number of iterations, but the results are slightly better than in the competitive case. But this improvement is not uniform over the three RL agents. SARSA is clearly benefited in this situation, and next Q-learning, while QV-learning being less able (or more altruistic) coping with the cooperation. This suggests that each algorithm may take different 'roles'. The difference between SARSA and Q-learning might be explained because SARSA takes more states and actions into account, and this may be beneficial when rewards are averaged.

## 7 Scenario measuring both competition and cooperation

Finally, we examine another scenario where we now have competition and cooperation at the same time, using the notion of 'team'. We define two teams, one with two Q-learning agents and the other one with two SARSA agents. Inside each team the rewards go to the same bag, but different teams compete for the rewards. This is shown in Fig. 5. In general, the results are poorer than with three agents in the cooperative case (Fig. 4, right). This can be explained because here we have four agents instead of three, but also because having two teams is a more complex scenario than having just one team.

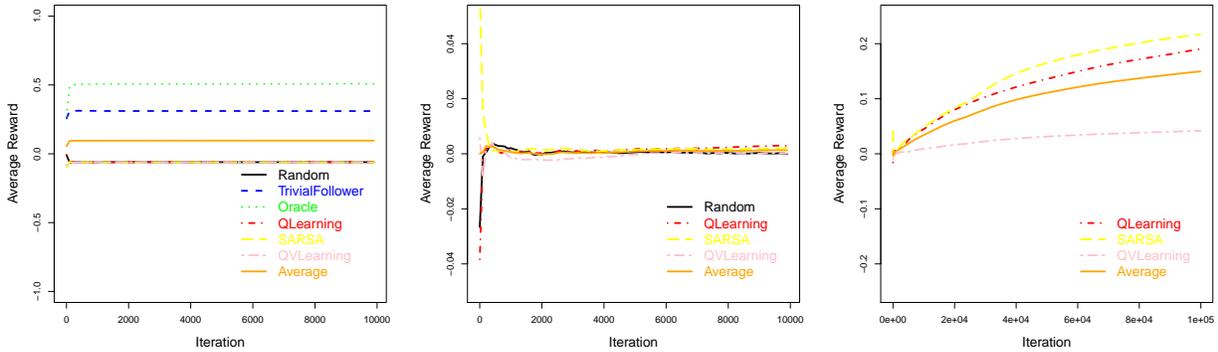

Figure 4: Cooperative scenarios. Left: The 6 evaluated agents. Middle: The 3 RL agents along with the random agent. Right: The 3 RL agents without the random agent. 100 environments each.

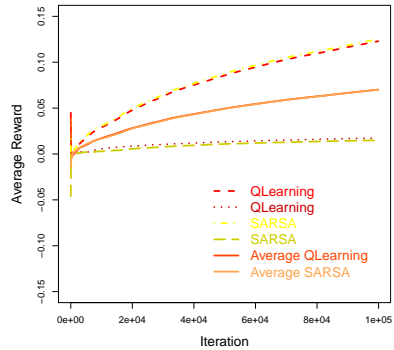

Figure 5: Two teams scenario. One team with two Q-learning agents against another team with two SARSA agents. Results with 100 environments.

The results show that there are no significant differences between both teams. However, there are important differences between the two components of each team. This can be observed in Fig. 5, where we assign the best results in the team to the first entry and the worst results to the second entry. That means that the plot just shows the difference in (average) performance between the best component in the team and the worst component in the team. We see that this difference is very significant. While there is usually an agent in the team which performs around 0.13, the other agent stays at a very low result close to 0. It is not clear which role this second agent takes.

## 8 Discussion

In the previous sections, we have analysed several scenarios. In them, a test which was originally designed to measure the intelligence of an agent against an environment without other agents is adapted to other scenarios where other agents are introduced in the environments. As expected, working with many agents makes things much more complex.

We see that performance can be seriously degraded by the inclusion of other agents with null intelligence, as a random agent. This is surprising if we look at this from the point of view of game theory (especially from the point of view of two-players games), but it is much more natural if we realise that it is more difficult to attain a goal if there is another agent bugging around (even randomly). This is extreme in the case of RL agents, because random agents can be considered noise, and this multiplies the state space.

All this means that the difficulty of a task is no longer related to the complexity of the environment in a tight way, as it was for the single agent situation. We can compare the complexity of the environment (excluding the evaluated agents) and the results for the scenarios where only the three RL agents are used, i.e. Fig. 2 (right) and Fig. 4 (right). In order to approximate the environment complexity, we use the size of a compressed coding of the concatenation of the space description $S$ and the description of the pattern for *Good* and *Evil*, denoted by $P$. More formally, we calculate an approximation to its (Kolmogorov) complexity, denoted by $K^{approx}$ as follows:

$$K^{approx} = LZ(S, P)$$

where $LZ$ is just the 'gzip' method given by the *memCompress* function in R, a GNU project implementation of Lempel-Ziv coding. The results of this comparison are shown in Fig. 6. We see that there is still a relation between the complexity of the environment and the result, while this relation is stronger for Qlearning and SARSA in the cooperative case. In fact, the results for Qlearning and SARSA are very good when the complexity is very low. This means that in very simple cases RL agents are able to perform well, even in social scenarios. This seems to suggest that the difficulty of a social environment is a cumulative issue, which sums up the complexity of the environment and the complexity/performance/noise of other agents.

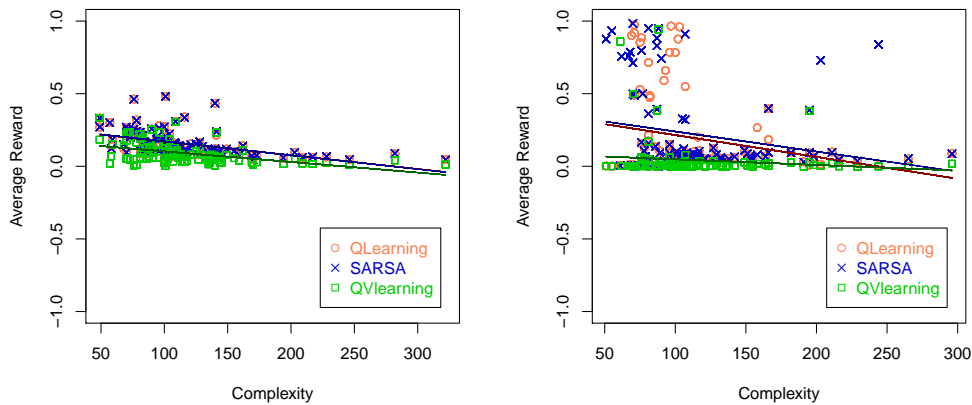

Figure 6: Relation between environment complexity and results of the three RL agents for the competitive case (Top) and the cooperative case (Bottom). Linear regression is also shown for each agent.

Another thing we can observe is the almost identical behaviour of Qlearning and SARSA in the competitive case, while they diverge in the cooperative case. This might be explained by the fact that in a cooperative situation, it makes sense if the agents go to different (non-optimal) cells, since their rewards are accumulated. If two agents go to the same cell, then they have to share the reward.

Finally, let us go back to the notion of the Darwin-Wallace distribution and see what lessons can be learnt from these experiments. One of the purported problems of this distribution is that it might take many iterations to reach a level where some social behaviour can be evaluated. We see that this may not be the case. For instance, we have seen that even using very simple agents, their mere introduction in an environment makes that the performance of other agents plunge. This means that even in just one generation (from $i = 0$ to $i = 1$) the Darwin-Wallace distribution makes a big leap. The intelligence of the base agents is not so relevant to make things much more difficult. In other words, the population of many agents, even dull agents, makes thing more difficult. However, there are still some differences in the way this difficulty is handled by different agents. In fact, apart from the oracle (and perhaps the trivial follower), the RL agents cannot be considered very intelligent. This suggests that the evaluation of social intelligence could possibly be performed against other agents of inferior degree of intelligence.

In fact, it would be extremely informative to repeat the experiment performed with humans and RL agents in [11] by using one of these simple multiagent environments. We guess that while humans will still be able to manage, the collapse that we observe in the RL agents would show that the mere introduction of some simple social behaviour may show the real differences between these two types of agents.

## 9 Conclusions

Measuring social intelligence without language seems a difficult issue. However, if we just define social intelligence as behaving well in multiagent environments, we 'only' need to solve which 'multiagent environments' should be chosen as test tasks. This is precisely what the Darwin-Wallace distribution aims at. However, there are many questions around the feasibility of such a proposal. In this paper, we have performed a series of experiments which may shed some light on these questions.

From here, we have pushed forward the idea of 'multiagent intelligence test', which is an intelligence test where there are other agents in the environments. This is a new notion, since the kind of intelligence tests we are used to are typically those where the evaluated agent has to solve some tasks or where it has to be interrogated by other agents (interviews, Turing test, etc.), but the other agents are not *inside* the test. The closest area is an old companion of artificial intelligence, *games*, especially multiplayer games, but it has only been recently proposed as a testbed for measuring intelligence [8]. However, the role of the opponent and its intelligence has not been clarified, especially if we want a test to give an absolute magnitude, not only comparing a pair of agents. So the notion of multiagent intelligence test where we can previously assess the intelligence of the other agents in a recursive fashion is an appealing idea.

However, before constructing a multiagent intelligence test based on these ideas, we need to better understand some phenomena that take place when we include other agents and let them compete and cooperate. Many other experiments about multiagent systems and the role of competition and cooperation have been made in the past, especially in the area of evolutionary game theory, but this paper has designed a set of experiments and has analysed the results from the point of view of how the inclusion of more agents may affect the base environment as a testbed for measuring (social) intelligence. Naturally, much more experiments must be done before embarking on the challenge of a true (and feasible) social intelligence test.

In this regard, the experiments shown in this paper could be extended in many ways. For instance, for the RL agents, we only consider model-free techniques whose search space grows geometrically as more agents are there. It would be interesting to see the results for model-based techniques or RL algorithms using function approximations, as well as other RL algorithms which work better when the Markov property does not hold (which is the general case in multiagent systems). Similarly, some other RL algorithms which are specialised for multiagent settings, such as Frequently Adjusted Q-learning [13] might give different results.

Other issues which could be reconsidered is the way we modify the reward system to make the test competitive or cooperative. In the latter case, e.g., instead of an arithmetic mean of rewards we could use the geometric mean or a harmonic mean (to prevent from getting 0 when one agent scores 0). This would make cooperation even more valuable, since weak agents would be taken care of by more powerful (intelligent) agents. But this goes beyond pure cooperation into the realm of selflessness. Another possible direction would be to remove the agents *Good* and *Evil* and let all the agents be able to generate rewards (and perhaps other objects) for the other agents. This would make things more difficult, but it would allow for more elaborate scenarios for competition, cooperation and communication.

In the end, we could say that it is important to see which questions multiagent learning is the answer to [19], but it is also important to see how much the technology of autonomous agents and multiagent systems is progressing in order to meet these expectations. Necessarily, we need measuring tools for that. But, interestingly, perhaps the answer to measuring social intelligence can also be found in multiagent systems.

## 10 Acknowledgements

We want to thank Hado van Hasselt for his RL algorithms library, which was useful when we were implementing our algorithms. This work was supported by the MEC projects EXPLORA-INGENIO TIN 2009-06078-E, CONSOLIDER-INGENIO 26706 and TIN 2010-21062-C02-02, and GVA project PROMETEO/2008/051. Javier Insa-Cabrera was sponsored by Spanish MEC-FPU grant AP2010-4389.